\documentclass[runningheads]{llncs}

\usepackage{accv}

\usepackage{accvabbrv}

\usepackage{graphicx}
\usepackage{booktabs}

\usepackage[accsupp]{axessibility}  

\usepackage{hyperref}

\usepackage{orcidlink}

\usepackage{multirow}
\usepackage{booktabs}
\usepackage{graphicx}
\begin{document}

\title{Enhanced Survival Prediction in Head and Neck
Cancer Using Convolutional Block Attention and
Multimodal Data Fusion} 

\titlerunning{Enhanced Survival Prediction in Head and Neck
Cancer}

\author{Aiman Farooq\inst{1}\orcidlink{0000-0002-3238-7962} \and
Utkarsh Sharma\inst{2}\orcidlink{0009-0005-4073-1099} \and
Deepak Mishra\inst{1}\orcidlink{0000-0002-4078-9400}}

\authorrunning{A.~Farooq et al.}

\institute{Indian Institute of Technology Jodhpur, Jodhpur, Rajasthan 342030, India \email{farooq.1@iitj.ac.in}\\ \and
Indian Institute of Science Education and Research Bhopal, Bhopal, Madhya Pradesh 462066, India
\email{lncs@springer.com}\\
}

\maketitle

\begingroup\renewcommand\thefootnote{
}
\footnotetext{A. Farooq and U. Sharma -- Equal contribution.}
\endgroup

\begin{abstract}
  Accurate survival prediction in head and neck cancer (HNC) is essential for guiding clinical decision-making and optimizing treatment strategies. Traditional models, such as Cox proportional hazards, have been widely used but are limited in their ability to handle complex multi-modal data. This paper proposes a deep learning-based approach leveraging CT and PET imaging modalities to predict survival outcomes in HNC patients. Our method integrates feature extraction with a Convolutional Block Attention Module (CBAM) and a multi-modal data fusion layer that combines imaging data to generate a compact feature representation. The final prediction is achieved through a fully parametric discrete-time survival model, allowing for flexible hazard functions that overcome the limitations of traditional survival models. We evaluated our approach using the HECKTOR and HEAD-NECK-RADIOMICS-HN1 datasets, demonstrating its superior performance compared to conventional statistical and machine learning models. The results indicate that our deep learning model significantly improves survival prediction accuracy, offering a robust tool for personalized treatment planning in HNC.

\keywords{Survival Prediction  \and Cancer \and Precision Medicine \and C-index.}
\end{abstract}

\section{Introduction}
\label{sec:intro}

Survival prediction is critical in the management and care of various diseases, particularly in the realm of cancer diagnosis. For patients diagnosed with cancers such as lung, breast, colorectal, or head and neck cancer, understanding survival probabilities helps guide crucial clinical decisions, from treatment planning to resource allocation. Head and neck cancer (HNC) ranks as the seventh most common cancer worldwide, with over 660,000 new cases and 325,000 deaths reported annually \cite{sung2021global}, and this number is expected to increase by 30\% annually by the year 2030 \cite{gormley2022reviewing}. Accurate survival estimates for HNC allow oncologists to tailor treatment strategies based on a patient’s prognosis, determining whether aggressive interventions like surgery, chemotherapy, or radiation are appropriate or if a palliative care approach would provide the best quality of life. These predictions are also essential for counseling patients and families, setting realistic expectations, and aligning care goals with the patient’s values and preferences. However, the task of survival prediction is complex and involves multiple data modalities, including clinical, imaging, and sometimes even genomic modeling. However, the major hindrance in this domain is the availability of data. Most datasets generally contain imaging modalities with clinical and other relevant data lacking for most patients. Achieving accurate and reliable predictions from imaging modalities alone is a significant and ongoing problem in this domain. 

Deep learning has revolutionized the medical domain, offering transformative solutions for some of the most complex and critical healthcare challenges. Survival prediction for HNC typically involves integrating data from various sources, including CT scans, PET scans, and clinical information. Features extracted from CT imaging are highly relevant in predicting survival outcomes in HNC \cite{gupta2021radiomics,kong2022deep,li2023prognostic}. PET imaging has also been extensively studied for its role in prognosis, with metabolic activity and other biomarkers from PET scans linked to survival predictions \cite{huang2021pet,wang2022integrating}. Additionally, molecular data have proven valuable, providing insights into the biological mechanisms driving cancer progression and helping to enhance the accuracy of survival models \cite{chen2019integrative}. Traditionally, survival prediction models such as Cox proportional hazards models \cite{xu2019nomogram}, random survival forests \cite{liu2022integrated}, and support vector machines \cite{wu2018development} have relied on handcrafted features derived from imaging or clinical data. However, with the increasing application of deep learning, advanced models based on convolutional neural networks (CNNs) \cite{o2015introduction} have been developed, surpassing traditional approaches. For instance, Hu et al. \cite{hu2020radiomics} employed CNNs to extract imaging features from CT scans and integrated them with clinical data for survival prediction in HNC patients.
Moreover, models like DeepSurv \cite{chen2021deepsurv} have demonstrated strong performance by directly learning from multi-modal data sources. Recognizing the limitations of relying solely on a single modality, several studies have proposed multi-modal models to capture complementary information from diverse data sources. Approaches such as those by Wang et al. \cite{wang2020deep} and Jin et al. \cite{jin2023fusion} have combined imaging and clinical for more accurate survival prediction in HNC. Additionally, models like DeepMM \cite{sun2021deepmm} have shown how multi-modal fusion can significantly improve performance by leveraging the strengths of each modality.

In exploring survival prediction in HNC, we focus on leveraging deep learning architectures specifically designed to integrate information from CT and PET imaging modalities. Our study's primary goal is to harness these imaging techniques' strengths to develop robust survival prediction models. CT scans provide crucial structural details, while PET scans offer metabolic insights—together, they capture complementary features critical for accurate prognosis. We employ advanced models like convolutional neural networks (CNNs) to extract and combine relevant features from these modalities, allowing for a comprehensive understanding of the tumor's behavior and progression. Although adding clinical and genomic data has improved survival predictions in prior studies, such data is often unavailable or incomplete for a significant portion of patients. Clinical and genomic information is typically available for only 10-20\% of HNC cases, limiting the practicality of relying on these datasets. We optimize predictions using  CT and PET imaging along with the available clinical data. 
\section{Related Work}
Survival prediction in HNC is a critical area of research that has been significantly advanced by using various statistical, machine learning (ML), and deep learning (DL) models. Traditionally, statistical models like the Cox proportional hazards model have been the backbone of survival analysis in clinical research. This model is particularly valued for its ability to assess the impact of various covariates on survival time. \cite{smith2012cox} employed the Cox model to evaluate the survival outcomes of HNC patients based on clinical and demographic factors. Their study achieved a concordance index of 0.71, highlighting the model’s utility in providing reasonably accurate survival predictions. Another widely used approach is the Kaplan-Meier estimator, which provides a non-parametric statistic for estimating the survival function from lifetime data. \cite{lee2014kaplan} utilized this method to analyze the survival probabilities of different HNC treatment groups, illustrating the method’s effectiveness in comparative survival analysis, especially in stratifying patient groups.

With the growing complexity of patient data and the need for more accurate predictions, machine learning models have increasingly been employed in HNC survival analysis. Random forests (RF) \cite{breiman2001random} and support vector machines (SVM) \cite{cortes1995support} are the most commonly used ML techniques. RF models, known for their robustness in handling large datasets and reducing overfitting through ensemble learning, have been effectively used for survival prediction. For instance, \cite{li2017random} applied RF to predict survival outcomes in HNC patients, achieving an area under the curve (AUC) of 0.79, underscoring the model's ability to handle complex datasets with numerous variables. Similarly, \cite{wang2016svm} used SVMs to predict patient survival based on gene expression profiles, achieving a classification accuracy of 83\%, demonstrating the SVM's effectiveness in high-dimensional data settings. Furthermore, artificial neural networks (ANNs) have been utilized to capture non-linear relationships between covariates and survival time. \cite{gupta2018ann} reported a significant improvement in predictive accuracy using ANNs compared to traditional Cox models, with their model achieving a concordance index of 0.76.

The advent of deep learning has introduced even more sophisticated approaches to survival prediction in HNC. Convolutional neural networks (CNNs), traditionally used for image processing, have been adapted to process complex multi-dimensional data in survival analysis. For example, \cite{liu2019cnn} used CNNs to integrate imaging data with clinical variables, achieving a concordance index of 0.80 in predicting overall survival, a marked improvement over traditional ML models. \cite{patel2020rnn} employed RNNs to model time-to-event data in HNC, finding that their approach could predict survival with a mean absolute error of 3.1 months, thereby offering a promising tool for dynamic survival prediction as more patient data becomes available over time.

 In a study by \cite{sun2021radiogenomics}, a deep learning model combining CNNs with radiomic and genomic data was used to predict survival outcomes in HNC patients, achieving an AUC of 0.85. Integrating diverse data types represents a significant step in pursuing personalized medicine, enabling more tailored and accurate predictions for individual patients. Vale et al. \cite{vale2021long} proposed MultiSurv to predict long-term cancer survival by integrating imaging ( whole slide imaging) and clinical data for a pan-cancer approach. Katzman et al. \cite{katzman2018deepsurv} proposed a deep learning-based survival model, DeepSurv, for breast cancer. It extends the Cox proportional hazards model and utilizes neural networks to model complex, nonlinear relationships in survival data.

Overall, applying statistical, machine learning, and deep learning models in survival prediction for head and neck cancer has evolved significantly. Traditional statistical models continue to provide foundational insights while incorporating ML and DL techniques, which have markedly improved the accuracy and applicability of survival predictions. These advancements are paving the way for more personalized and effective treatment strategies, ultimately aiming to improve patient outcomes in this challenging area of oncology.
\section{Methodology}
\subsection{Dataset}

We have used two main head and neck cancer datasets for this study, which include the Head \& neCK TumOR segmentation and outcome prediction (HECKTOR) data set \cite{oreiller2022head} and the HEAD-NECK-RADIOMICS-HN1 collection \cite{aerts2014,wee2019}. The HECKTOR dataset is a multi-modal, multi-center collection specifically curated for head and neck cancer research. It comprises PET and CT scans from 488 patients and corresponding tumor segmentation masks. This dataset, sourced from seven different centers, provides a rich resource for developing and validating segmentation algorithms and outcome prediction models. The dataset also includes detailed clinical data, such as Recurrence-Free Survival (RFS) information, encompassing time-to-event data and censoring status. The HEAD-NECK-RADIOMICS-HN1 collection is sourced from The Cancer Imaging Archive (TCIA). It includes imaging and clinical data from 137 patients suffering from head-and-neck squamous cell carcinoma (HNSCC) patients treated at MAASTRO Clinic, The Netherlands. CT scans, manual delineations, and clinical and survival data are available for these patients. PET images in the dataset have pixel sizes ranging from 1.95 mm to 5.47 mm, slice thicknesses from 2.02 mm to 5 mm, and matrix sizes between 128x128 and 256x256 pixels. CT images feature pixel sizes ranging from 0.68 mm to 1.95 mm, slice thicknesses from 1.5 mm to 5 mm, and a matrix size of 512x512 pixels. 

\subsection{Network Architecture}
\begin{figure}
    \centering
    \includegraphics[scale=0.22]{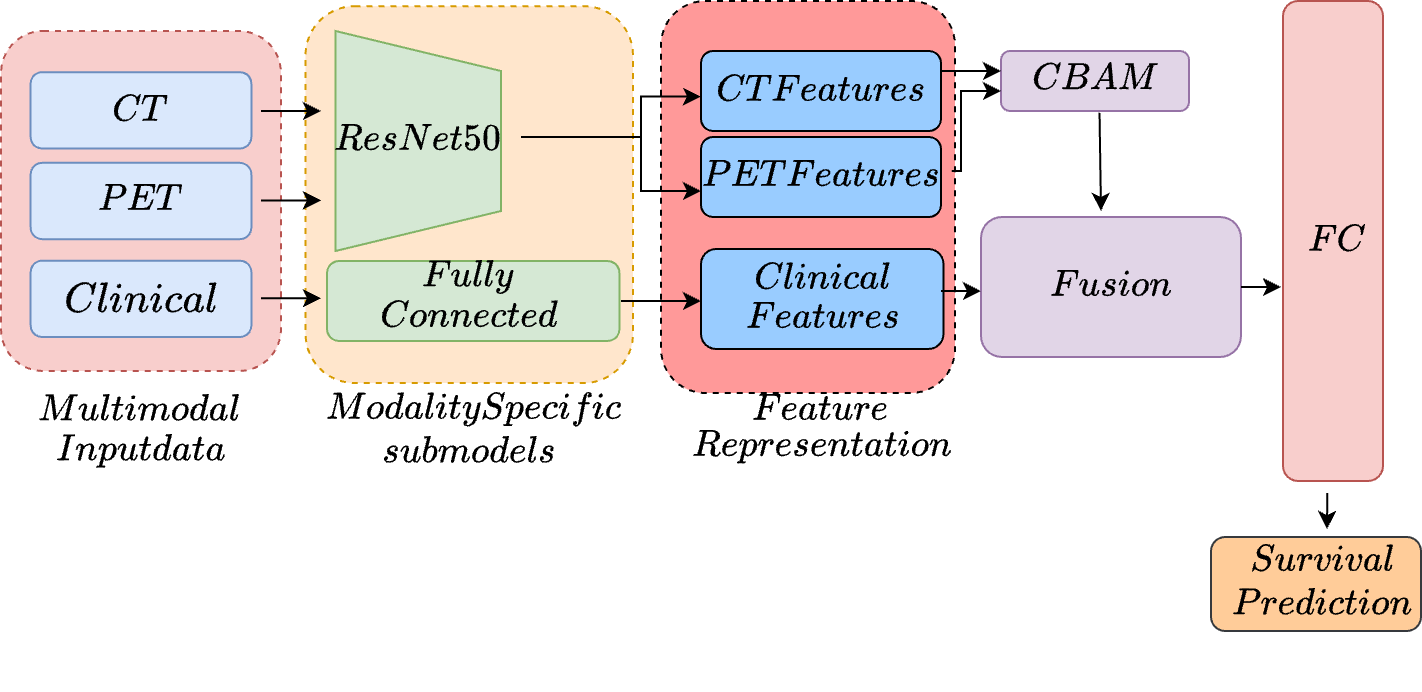}
    \caption{Proposed multi-modal deep learning model for survival prediction, integrating CT, PET, and clinical data with feature fusion and CBAM for enhanced prediction accuracy. Final survival predictions are made using a fully connected layer.}
    \label{fig:label1}
\end{figure}

The overall architecture of the proposed model, as shown in Fig. \ref{fig:label1}, is structured around four core modules: a feature representation module, an attention module, a multi-modal data fusion layer, and an output submodel. These components integrate and process data from multiple modalities—CT images, PET images, and clinical data—to predict conditional survival probabilities for discrete follow-up time intervals. 
The feature representation module is designed with dedicated submodels for each data modality. The submodels are based on a ResNet-50 3D architecture \cite{he2015deep} for the CT and PET images. This deep convolutional neural network is optimized explicitly for handling volumetric data, such as medical images. The architecture of ResNet-50 3D comprises multiple layers of 3D convolutional filters, along with batch normalization, ReLU activation functions, and max pooling layers. These layers capture spatial hierarchies and intricate local textures within the CT and PET scans. The output from this stage is a collection of detailed feature maps for each imaging type, capturing important spatial details needed for further analysis. 
In parallel, the clinical data is passed through a fully connected (FC) layer, which transforms the input features into a dense representation. This dense representation captures the complex relationships between the clinical variables, facilitating their integration with the image-based features extracted from the CT and PET data.

After feature extraction, a Convolutional Block Attention Module (CBAM)\cite{woo2018cbam} is applied to the feature maps. CBAM enhances the discriminative power of the extracted features through two sequential stages: channel attention and spatial attention. The channel attention mechanism emphasizes the most relevant feature maps, answering the question of "what" is essential for the prediction task. Following this, spatial attention refines the focus further by determining "where" the network should concentrate its attention on the image. By applying this dual attention mechanism, CBAM selectively enhances the most informative aspects of the CT and PET data, leading to more refined and relevant feature maps.

Following feature extraction, the outputs from these submodels are integrated within the multi-modal data fusion layer. This layer reduces the set of feature representation vectors into a single, compact fusion vector, which serves as the input to the subsequent output submodel. Formally, let $ Z = [z_1, \dots, z_n]$ denote the matrix composed of the feature representation vectors from the different modalities, where each vector $z_l \in \mathbb{R}^m$ represents the features extracted from the $l^{\text{th}}$ modality. The fusion vector $c \in \mathbb{R}^m $ is then computed as the row-wise maximum of $Z$, with the $ k^{\text{th}}$ element of $c$ given by:

\begin{equation}
    c_k = \max_{1 \leq l \leq n} z_{k,l}, \quad k = 1, \dots, m.
\end{equation}

This approach results in a fused feature representation corresponding to the maxima across the different data modalities.

The final module of the proposed model architecture is the prediction submodel, which is designed as a fully parametric discrete-time survival model parameterized by a deep neural network. The model is trained using stochastic gradient descent (SGD), with the assumption that follow-up time is discrete, represented as a set of time intervals \(\{t_1, t_2, \dots, t_p\}\), where each \(t_j\) marks the upper limit of the \(j^{\text{th}}\) interval.

For each study subject, the hazard function \(h_j\) defines the probability that the event of interest occurs within interval \(j\), conditional on survival up to the beginning of the interval. The model is trained by minimizing the negative log-likelihood of the observed data. Specifically, for each time interval \(j\), the log-likelihood is computed as:
\begin{equation}
    \sum_{i=1}^a {d_j} \log(h_j^{(i)}) + \sum_{i=d_j+1}^{r_j} \log(1 - h_j^{(i)}),
\end{equation}

where \(h_j^{(i)}\) is the hazard probability for the \(i^{\text{th}}\) subject during interval \(j\). The total loss is the sum of the negative log-likelihoods across all time intervals. The first term in the log-likelihood encourages the model to increase the hazard rate \(h_j\) for subjects who experience the event within interval \(j\). In contrast, the second term encourages the model to increase the predicted survival probability \(1 - h_j\) for subjects who survive beyond the interval, including those censored.
In this implementation, the output layer of the prediction submodel consists of \(p\) units, each corresponding to one-time interval. A sigmoid activation function converts the output of each unit into the predicted conditional probability of surviving the respective interval, which is the complement of the conditional hazard rate \(1 - h_j\).

The predicted probability of a subject \(i\) surviving through the end of interval \(j\) is given by:

\begin{equation}
    S_j^{(i)} = \prod_{q=1}^{j} (1 - h_q^{(i)}).
\end{equation}

The loss function is a reformulation of the negative log-likelihood divided by the number of study subjects, which facilitates training with mini-batches of patients. 

This prediction submodel, integrated with the compact fusion vector generated by the multi-modal data fusion layer, enables the model to effectively translate the combined features into survival probability predictions, leveraging uncensored and censored data.

\section{Experiments and Results}
A uniform framework was established to provide a standardized testing environment for all models, ensuring consistency and fairness across various aspects. The proposed models were implemented using PyTorch. The models are evaluated using the widely popular Concordance index (Ctd) \cite{antolini2005time}, an extension of the widely used Harrell’s concordance index (C-index)\cite{harrell1982evaluating}.

The models were trained for 50 epochs, using a batch size of 4 to optimize computational efficiency and stability during training. The validation set was used to assess the performance during iterative model development. We trained the models using Adam stochastic gradient descent optimization. The learning rate was set to $2 \times 10^{-3}$, a value chosen based on preliminary experiments to ensure a balance between convergence speed and model performance. The dataset was divided into training, validation, and test sets with an 80-10-10 split.

\begin{table*}[htbp]
\centering
\caption{Evaluation of the prediction results  using the HEAD-NECK-RADIOMICS-HN1  dataset}
\label{tab:table2}
\begin{tabular}{ccc}
\toprule
Model & Modality & Ctd-index \\ \hline

 XGBoost \cite{huang2020artificial} & CT, PET & 0.5742 \\
 \midrule
 RF \cite{he2022artificial} & CT, PET & 0.5909 \\
 \midrule
 DeepSurv \cite{katzman2018deepsurv} & CT, PET & 0.6743 \\
 \midrule
MultiSurv \cite{vale2021long} & CT, PET & 0.7018 \\ 
  \midrule
  MultiSurv (RNC)\cite{zha2024rank} & CT, PET & 0.6811 \\ 
  \midrule

Ours & CT, PET  & \textbf{0.7272} \\ \bottomrule
\end{tabular}
\end{table*}
%
\begin{table*}[htbp]
\centering
\caption{Evaluation of the prediction results using the HECKTOR dataset}
\label{tab:table1}
\begin{tabular}{ccc}
\toprule
Model & Modality & Ctd-index \\ \hline

 XGBoost \cite{huang2020artificial} & CT, PET & 0.5810 \\
 \midrule
RF  \cite{he2022artificial} & CT \& PET & 0.6015 \\ 
 \midrule
 
\multicolumn{1}{l}{\multirow{2}{*}{Multisurv\cite{vale2021long}}}  & CT, PET & 0.6722  \\
 & CT, PET \& Clinical & 0.6489  \\ 
 \midrule
  \multicolumn{1}{l}{\multirow{2}{*}{Multisurv(RNC)\cite{zha2024rank}}}  & CT, PET & 0.6292  \\
 & CT, PET \& Clinical & 0.6214  \\ 
 \midrule
 

Ours & CT, PET  & \textbf{0.7010} \\ \bottomrule
\end{tabular}
\end{table*}

As shown in Table \ref{tab:table2}, initial experiments focused on integrating CT and PET images for survival prediction on the HN1 dataset. However, upon incorporating clinical data into the model, we do not observe any improvement in performance metrics. Despite the anticipated complementary nature of clinical data in enhancing predictive accuracy, the gains were insufficient to justify the added complexity. We also experimented with the Rank Consistency Loss (RNC Loss) \cite{zha2024rank}, designed to penalize incorrect ranking of survival risks among patients. Still, this approach did not yield meaningful improvements in our results. We observed a significant performance improvement by incorporating the attention module and explicitly targeting the feature maps generated from CT and PET images. This enhancement led to a noticeable increase in the ctd-index across all evaluated time intervals, highlighting the effectiveness of the attention mechanism in refining survival predictions. Table \ref{tab:table1} shows the results for the HECKTOR dataset, and we observe the same pattern here and see that using the CT and PET imaging features alone provides a considerable improvement in the performance while incorporating the clinical information leads to a drop in the performance.
\section{Conclusion}
In this study, we proposed a novel deep learning-based approach for survival prediction in head and neck cancer (HNC) that effectively integrates CT and PET imaging modalities. Our method leverages a Convolutional Block Attention Module (CBAM) for enhanced feature extraction and a multi-modal data fusion layer to combine imaging data into a compact representation. A fully parametric discrete-time survival model then utilizes this representation to predict survival probabilities across discrete time intervals. Our experimental results, evaluated on the HECKTOR and HEAD-NECK-RADIOMICS-HN1 datasets, demonstrate that our approach significantly outperforms traditional statistical models and other contemporary machine-learning techniques. Specifically, our method achieved superior C-index scores, highlighting its ability to provide more accurate survival predictions compared to existing methods.

The integration of CT and PET imaging modalities, coupled with advanced deep learning techniques, addresses some limitations of relying solely on a single data modality. By focusing on imaging data alone, our model provides a scalable and practical solution for survival prediction in real-world clinical settings, where clinical and genomic data might be incomplete or unavailable. Future work will explore incorporating additional data modalities and the potential benefits of further integrating clinical and genomic information to enhance predictive accuracy. 

Overall, our deep learning approach offers a robust and reliable tool for personalized treatment planning in HNC, contributing to the ongoing efforts to improve patient outcomes through advanced predictive modeling.

\bibliographystyle{splncs04}
\bibliography{main}
\end{document}